\documentclass[letterpaper, 10pt, journal, twoside]{IEEEtran}
\IEEEoverridecommandlockouts

\usepackage{cite}
\usepackage{amsmath,amssymb,amsfonts}
\usepackage{algorithmic}
\usepackage{graphicx}
\usepackage{textcomp}
\usepackage{xcolor}
\usepackage{colortbl}
\usepackage{listings}
\usepackage{tabularx}
\usepackage{booktabs}
\usepackage{siunitx}
\usepackage{standalone}
\usepackage[T1]{fontenc}
\usepackage{url}
\usepackage{flushend}
\usepackage{orcidlink}
\hypersetup{hidelinks}
\usepackage{pgfplots}
\usepackage{pgfplotstable}
\usepackage{xcolor}
\usepackage{booktabs}
\usepackage{array}
\pgfplotsset{compat=1.18}
\definecolor{bestgreen}{RGB}{38,132,82}
\definecolor{secondblue}{RGB}{52,101,164}
\definecolor{lightgray}{RGB}{190,190,190}
\definecolor{warnorange}{RGB}{210,118,30}
\definecolor{methodDlio}{RGB}{190,43,39}
\definecolor{methodEllipse}{RGB}{38,132,82}
\definecolor{methodFastlio}{RGB}{65,105,225}
\definecolor{methodFastlioQueue}{RGB}{98,139,232}
\definecolor{methodFastlioStrided}{RGB}{34,83,191}
\definecolor{methodIglio}{RGB}{101,84,164}
\definecolor{methodKissicp}{RGB}{196,160,0}
\definecolor{methodPelorus}{RGB}{204,71,150}
\definecolor{methodPelorusCuda}{RGB}{181,36,132}
\definecolor{methodPelorusEff}{RGB}{235,105,180}
\definecolor{methodPelorusHalf}{RGB}{242,143,199}
\definecolor{methodPelorusMulti2}{RGB}{146,43,180}
\definecolor{methodPelorusQueue}{RGB}{161,45,116}
\definecolor{methodRko}{RGB}{206,92,0}
\definecolor{methodSurfel}{RGB}{85,87,83}
\definecolor{loadGreen}{RGB}{44,162,95}
\definecolor{loadYellow}{RGB}{255,237,111}
\definecolor{loadRed}{RGB}{215,48,39}
\definecolor{turbo000}{RGB}{35,23,27}
\definecolor{turbo001}{RGB}{39,26,40}
\definecolor{turbo002}{RGB}{43,28,52}
\definecolor{turbo003}{RGB}{47,30,63}
\definecolor{turbo004}{RGB}{51,32,74}
\definecolor{turbo005}{RGB}{54,35,85}
\definecolor{turbo006}{RGB}{57,37,95}
\definecolor{turbo007}{RGB}{59,40,105}
\definecolor{turbo008}{RGB}{62,42,114}
\definecolor{turbo009}{RGB}{64,44,123}
\definecolor{turbo010}{RGB}{66,47,132}
\definecolor{turbo011}{RGB}{68,49,140}
\definecolor{turbo012}{RGB}{69,52,148}
\definecolor{turbo013}{RGB}{71,55,155}
\definecolor{turbo014}{RGB}{72,57,162}
\definecolor{turbo015}{RGB}{73,60,169}
\definecolor{turbo016}{RGB}{73,62,175}
\definecolor{turbo017}{RGB}{74,65,181}
\definecolor{turbo018}{RGB}{74,68,187}
\definecolor{turbo019}{RGB}{75,70,193}
\definecolor{turbo020}{RGB}{75,73,198}
\definecolor{turbo021}{RGB}{75,76,203}
\definecolor{turbo022}{RGB}{75,79,207}
\definecolor{turbo023}{RGB}{74,81,211}
\definecolor{turbo024}{RGB}{74,84,215}
\definecolor{turbo025}{RGB}{74,87,219}
\definecolor{turbo026}{RGB}{73,89,223}
\definecolor{turbo027}{RGB}{73,92,226}
\definecolor{turbo028}{RGB}{72,95,229}
\definecolor{turbo029}{RGB}{71,98,232}
\definecolor{turbo030}{RGB}{70,101,234}
\definecolor{turbo031}{RGB}{69,103,237}
\definecolor{turbo032}{RGB}{68,106,239}
\definecolor{turbo033}{RGB}{67,109,240}
\definecolor{turbo034}{RGB}{66,112,242}
\definecolor{turbo035}{RGB}{65,114,244}
\definecolor{turbo036}{RGB}{64,117,245}
\definecolor{turbo037}{RGB}{63,120,246}
\definecolor{turbo038}{RGB}{62,123,247}
\definecolor{turbo039}{RGB}{61,125,248}
\definecolor{turbo040}{RGB}{59,128,248}
\definecolor{turbo041}{RGB}{58,131,249}
\definecolor{turbo042}{RGB}{57,134,249}
\definecolor{turbo043}{RGB}{56,136,249}
\definecolor{turbo044}{RGB}{55,139,249}
\definecolor{turbo045}{RGB}{53,142,249}
\definecolor{turbo046}{RGB}{52,144,248}
\definecolor{turbo047}{RGB}{51,147,248}
\definecolor{turbo048}{RGB}{50,150,247}
\definecolor{turbo049}{RGB}{49,152,246}
\definecolor{turbo050}{RGB}{48,155,246}
\definecolor{turbo051}{RGB}{47,158,245}
\definecolor{turbo052}{RGB}{46,160,244}
\definecolor{turbo053}{RGB}{45,163,242}
\definecolor{turbo054}{RGB}{44,165,241}
\definecolor{turbo055}{RGB}{43,168,240}
\definecolor{turbo056}{RGB}{42,170,238}
\definecolor{turbo057}{RGB}{42,173,237}
\definecolor{turbo058}{RGB}{41,175,235}
\definecolor{turbo059}{RGB}{40,178,234}
\definecolor{turbo060}{RGB}{40,180,232}
\definecolor{turbo061}{RGB}{39,182,230}
\definecolor{turbo062}{RGB}{39,185,228}
\definecolor{turbo063}{RGB}{38,187,226}
\definecolor{turbo064}{RGB}{38,189,224}
\definecolor{turbo065}{RGB}{37,192,222}
\definecolor{turbo066}{RGB}{37,194,220}
\definecolor{turbo067}{RGB}{37,196,218}
\definecolor{turbo068}{RGB}{37,198,215}
\definecolor{turbo069}{RGB}{37,200,213}
\definecolor{turbo070}{RGB}{37,202,211}
\definecolor{turbo071}{RGB}{37,205,209}
\definecolor{turbo072}{RGB}{37,207,206}
\definecolor{turbo073}{RGB}{38,209,204}
\definecolor{turbo074}{RGB}{38,210,201}
\definecolor{turbo075}{RGB}{38,212,199}
\definecolor{turbo076}{RGB}{39,214,196}
\definecolor{turbo077}{RGB}{39,216,194}
\definecolor{turbo078}{RGB}{40,218,191}
\definecolor{turbo079}{RGB}{41,220,189}
\definecolor{turbo080}{RGB}{42,221,186}
\definecolor{turbo081}{RGB}{43,223,184}
\definecolor{turbo082}{RGB}{44,225,181}
\definecolor{turbo083}{RGB}{45,226,178}
\definecolor{turbo084}{RGB}{46,228,176}
\definecolor{turbo085}{RGB}{47,229,173}
\definecolor{turbo086}{RGB}{48,231,171}
\definecolor{turbo087}{RGB}{49,232,168}
\definecolor{turbo088}{RGB}{51,234,166}
\definecolor{turbo089}{RGB}{52,235,163}
\definecolor{turbo090}{RGB}{54,236,160}
\definecolor{turbo091}{RGB}{55,238,158}
\definecolor{turbo092}{RGB}{57,239,155}
\definecolor{turbo093}{RGB}{59,240,153}
\definecolor{turbo094}{RGB}{61,241,150}
\definecolor{turbo095}{RGB}{63,242,148}
\definecolor{turbo096}{RGB}{65,243,145}
\definecolor{turbo097}{RGB}{67,244,143}
\definecolor{turbo098}{RGB}{69,245,140}
\definecolor{turbo099}{RGB}{71,246,138}
\definecolor{turbo100}{RGB}{73,247,135}
\definecolor{turbo101}{RGB}{75,248,133}
\definecolor{turbo102}{RGB}{78,249,131}
\definecolor{turbo103}{RGB}{80,249,128}
\definecolor{turbo104}{RGB}{82,250,126}
\definecolor{turbo105}{RGB}{85,250,124}
\definecolor{turbo106}{RGB}{87,251,121}
\definecolor{turbo107}{RGB}{90,251,119}
\definecolor{turbo108}{RGB}{93,252,117}
\definecolor{turbo109}{RGB}{95,252,115}
\definecolor{turbo110}{RGB}{98,253,113}
\definecolor{turbo111}{RGB}{101,253,110}
\definecolor{turbo112}{RGB}{104,253,108}
\definecolor{turbo113}{RGB}{106,253,106}
\definecolor{turbo114}{RGB}{109,254,104}
\definecolor{turbo115}{RGB}{112,254,102}
\definecolor{turbo116}{RGB}{115,254,100}
\definecolor{turbo117}{RGB}{118,254,98}
\definecolor{turbo118}{RGB}{121,254,96}
\definecolor{turbo119}{RGB}{124,253,94}
\definecolor{turbo120}{RGB}{127,253,93}
\definecolor{turbo121}{RGB}{130,253,91}
\definecolor{turbo122}{RGB}{133,253,89}
\definecolor{turbo123}{RGB}{136,252,87}
\definecolor{turbo124}{RGB}{139,252,86}
\definecolor{turbo125}{RGB}{142,252,84}
\definecolor{turbo126}{RGB}{145,251,82}
\definecolor{turbo127}{RGB}{149,251,81}
\definecolor{turbo128}{RGB}{152,250,79}
\definecolor{turbo129}{RGB}{155,249,78}
\definecolor{turbo130}{RGB}{158,249,76}
\definecolor{turbo131}{RGB}{161,248,75}
\definecolor{turbo132}{RGB}{164,247,73}
\definecolor{turbo133}{RGB}{167,246,72}
\definecolor{turbo134}{RGB}{170,246,70}
\definecolor{turbo135}{RGB}{173,245,69}
\definecolor{turbo136}{RGB}{176,244,68}
\definecolor{turbo137}{RGB}{179,243,66}
\definecolor{turbo138}{RGB}{182,242,65}
\definecolor{turbo139}{RGB}{185,240,64}
\definecolor{turbo140}{RGB}{188,239,63}
\definecolor{turbo141}{RGB}{191,238,62}
\definecolor{turbo142}{RGB}{194,237,60}
\definecolor{turbo143}{RGB}{197,235,59}
\definecolor{turbo144}{RGB}{200,234,58}
\definecolor{turbo145}{RGB}{203,233,57}
\definecolor{turbo146}{RGB}{205,231,56}
\definecolor{turbo147}{RGB}{208,230,55}
\definecolor{turbo148}{RGB}{211,228,54}
\definecolor{turbo149}{RGB}{213,227,53}
\definecolor{turbo150}{RGB}{216,225,52}
\definecolor{turbo151}{RGB}{219,223,52}
\definecolor{turbo152}{RGB}{221,222,51}
\definecolor{turbo153}{RGB}{223,220,50}
\definecolor{turbo154}{RGB}{226,218,49}
\definecolor{turbo155}{RGB}{228,216,48}
\definecolor{turbo156}{RGB}{230,214,48}
\definecolor{turbo157}{RGB}{233,212,47}
\definecolor{turbo158}{RGB}{235,210,46}
\definecolor{turbo159}{RGB}{237,208,45}
\definecolor{turbo160}{RGB}{239,206,45}
\definecolor{turbo161}{RGB}{241,204,44}
\definecolor{turbo162}{RGB}{243,202,43}
\definecolor{turbo163}{RGB}{244,200,43}
\definecolor{turbo164}{RGB}{246,198,42}
\definecolor{turbo165}{RGB}{248,196,42}
\definecolor{turbo166}{RGB}{249,193,41}
\definecolor{turbo167}{RGB}{251,191,40}
\definecolor{turbo168}{RGB}{252,189,40}
\definecolor{turbo169}{RGB}{253,186,39}
\definecolor{turbo170}{RGB}{255,184,39}
\definecolor{turbo171}{RGB}{255,181,38}
\definecolor{turbo172}{RGB}{255,179,38}
\definecolor{turbo173}{RGB}{255,177,37}
\definecolor{turbo174}{RGB}{255,174,37}
\definecolor{turbo175}{RGB}{255,172,36}
\definecolor{turbo176}{RGB}{255,169,36}
\definecolor{turbo177}{RGB}{255,166,35}
\definecolor{turbo178}{RGB}{255,164,35}
\definecolor{turbo179}{RGB}{255,161,34}
\definecolor{turbo180}{RGB}{255,159,34}
\definecolor{turbo181}{RGB}{255,156,34}
\definecolor{turbo182}{RGB}{255,153,33}
\definecolor{turbo183}{RGB}{255,151,33}
\definecolor{turbo184}{RGB}{255,148,32}
\definecolor{turbo185}{RGB}{255,145,32}
\definecolor{turbo186}{RGB}{255,142,31}
\definecolor{turbo187}{RGB}{255,140,31}
\definecolor{turbo188}{RGB}{255,137,30}
\definecolor{turbo189}{RGB}{255,134,30}
\definecolor{turbo190}{RGB}{255,131,30}
\definecolor{turbo191}{RGB}{255,129,29}
\definecolor{turbo192}{RGB}{255,126,29}
\definecolor{turbo193}{RGB}{255,123,28}
\definecolor{turbo194}{RGB}{255,120,28}
\definecolor{turbo195}{RGB}{255,117,27}
\definecolor{turbo196}{RGB}{255,115,27}
\definecolor{turbo197}{RGB}{255,112,26}
\definecolor{turbo198}{RGB}{254,109,26}
\definecolor{turbo199}{RGB}{252,106,26}
\definecolor{turbo200}{RGB}{251,104,25}
\definecolor{turbo201}{RGB}{249,101,25}
\definecolor{turbo202}{RGB}{248,98,24}
\definecolor{turbo203}{RGB}{246,95,24}
\definecolor{turbo204}{RGB}{244,92,23}
\definecolor{turbo205}{RGB}{243,90,23}
\definecolor{turbo206}{RGB}{241,87,22}
\definecolor{turbo207}{RGB}{239,84,22}
\definecolor{turbo208}{RGB}{237,82,21}
\definecolor{turbo209}{RGB}{235,79,20}
\definecolor{turbo210}{RGB}{233,76,20}
\definecolor{turbo211}{RGB}{230,74,19}
\definecolor{turbo212}{RGB}{228,71,19}
\definecolor{turbo213}{RGB}{226,69,18}
\definecolor{turbo214}{RGB}{224,66,18}
\definecolor{turbo215}{RGB}{221,64,17}
\definecolor{turbo216}{RGB}{219,61,16}
\definecolor{turbo217}{RGB}{216,59,16}
\definecolor{turbo218}{RGB}{214,56,15}
\definecolor{turbo219}{RGB}{211,54,15}
\definecolor{turbo220}{RGB}{209,52,14}
\definecolor{turbo221}{RGB}{206,49,13}
\definecolor{turbo222}{RGB}{203,47,13}
\definecolor{turbo223}{RGB}{201,45,12}
\definecolor{turbo224}{RGB}{198,43,11}
\definecolor{turbo225}{RGB}{196,41,11}
\definecolor{turbo226}{RGB}{193,39,10}
\definecolor{turbo227}{RGB}{190,37,10}
\definecolor{turbo228}{RGB}{188,35,9}
\definecolor{turbo229}{RGB}{185,33,8}
\definecolor{turbo230}{RGB}{183,31,8}
\definecolor{turbo231}{RGB}{180,29,7}
\definecolor{turbo232}{RGB}{177,28,6}
\definecolor{turbo233}{RGB}{175,26,6}
\definecolor{turbo234}{RGB}{172,24,5}
\definecolor{turbo235}{RGB}{170,23,4}
\definecolor{turbo236}{RGB}{168,22,4}
\definecolor{turbo237}{RGB}{165,20,3}
\definecolor{turbo238}{RGB}{163,19,2}
\definecolor{turbo239}{RGB}{161,18,2}
\definecolor{turbo240}{RGB}{159,17,1}
\definecolor{turbo241}{RGB}{157,16,0}
\definecolor{turbo242}{RGB}{155,15,0}
\definecolor{turbo243}{RGB}{154,14,0}
\definecolor{turbo244}{RGB}{152,14,0}
\definecolor{turbo245}{RGB}{150,13,0}
\definecolor{turbo246}{RGB}{149,12,0}
\definecolor{turbo247}{RGB}{148,12,0}
\definecolor{turbo248}{RGB}{147,12,0}
\definecolor{turbo249}{RGB}{146,12,0}
\definecolor{turbo250}{RGB}{145,11,0}
\definecolor{turbo251}{RGB}{145,12,0}
\definecolor{turbo252}{RGB}{144,12,0}
\definecolor{turbo253}{RGB}{144,12,0}
\definecolor{turbo254}{RGB}{144,12,0}
\definecolor{turbo255}{RGB}{144,13,0}
\pgfplotsset{colormap={resourceGreenRed}{rgb255=(44,162,95) rgb255=(255,237,111) rgb255=(215,48,39)}}

\DeclareSIUnit[quantity-product={}, range-units=repeat]\percent{\%}
\def\BibTeX{{\rm B\kern-.05em{\sc i\kern-.025em b}\kern-.08em
    T\kern-.1667em\lower.7ex\hbox{E}\kern-.125emX}}

\definecolor{bestcolor}{HTML}{990000}
\definecolor{mutedgrey}{gray}{0.6}

\newcommand{\secref}[1]{Section~\ref{sec:#1}}

\newcommand{\figref}[1]{Fig.~\ref{fig:#1}}

\newcommand{\best}[1]{\textbf{#1}}
\newcommand{\italtxt}[1]{\textcolor{black!50}{#1}}
\newcommand{\italbest}[1]{\italtxt{#1}}
\newcommand{\second}[1]{\underline{#1}}
\newcommand{\italsecond}[1]{\italtxt{#1}}

\newcommand{\DNF}{\multicolumn{1}{c}{d.n.f.}}
\newcommand{\DNFit}{\multicolumn{1}{c}{\italtxt{d.n.f.}}}

\newcolumntype{Y}{>{\centering\arraybackslash\S\hfil}X<{\hfil}}

\ifdefined\IEEEkeywords\else
\newenvironment{IEEEkeywords}{\raggedright\noindent\textbf{\textit{Index Terms}---}}{\par}
\fi

\newcommand{\coderepository}{\url{https://codeberg.org/stelzo/pelorus}}
\newcommand{\playbacktool}{the minot bag player}
\newcommand{\playbacktoolnote}{\footnote{\url{https://codeberg.org/stelzo/minot}}}

\begin{document}

\title{Pelorus: Resource-Aware Lidar-Inertial Odometry}

\author{Charlotte Sieh\,\orcidlink{0000-0001-7739-8844} and Mario Porrmann\,\orcidlink{0000-0003-1005-5753}%
\thanks{This work has been submitted to the IEEE for possible publication. Copyright may be transferred without notice, after which this version may no longer be accessible.}%
\thanks{This work was supported in part by the Lower Saxony Ministry of Science and Culture (MWK) through the zukunft.niedersachsen program of the Volkswagen Foundation under Project HybrInt (ZN4219). \textit{(Corresponding author: Charlotte Sieh.)}}%
\thanks{The authors are with the Computer Engineering Group, Osnabrück University, 49090 Osnabrück, Germany (e-mail: chsieh@uos.de; mporrmann@uos.de).}%
}
\markboth{Preprint submitted to IEEE Robotics and Automation Letters}
{Sieh and Porrmann: Pelorus: Resource-Aware Lidar-Inertial Odometry}

\maketitle
\raggedbottom

\begin{abstract}
Lidar-inertial odometry (LIO) has an efficiency problem. High-resolution lidars produce millions of points per second that must be processed online on the resource-constrained hardware of robots, cars, or drones, yet most LIO methods neither bound their memory nor report their compute footprint. We introduce Pelorus, a resource-aware LIO with a fixed-capacity multiresolution map of voxelized normal distributions and an iterated equivariant state estimator. A relevance-based eviction keeps the map memory bounded, and a maturity-weighted map update suppresses inconsistent observations. The pipeline runs in parallel on CPUs and GPUs. On five sequences from three public datasets, Pelorus achieves the lowest relative pose error in seven of ten metrics with a small, bounded memory footprint and, with GPU offloading, the lowest CPU load of all compared methods. On an NVIDIA Jetson Orin NX, it processes a 128-beam lidar at 10~Hz. Pelorus is open source with CUDA, Vulkan, and Metal backends.
\end{abstract}


\section{Introduction}
Mobile robots need reliable localization in unseen environments, and depending on external infrastructure such as the Global Positioning System (GPS) is a risk for full autonomy. Lidar-inertial odometry (LIO), which fuses lidar point clouds with measurements of an inertial measurement unit (IMU), has therefore become a standard for self-contained localization and the frontend of many simultaneous localization and mapping (SLAM) systems.

Since LIO runs on the mobile platform next to the tasks that rely on it, resource efficiency is a strict requirement. A method whose memory grows with the trajectory eventually fails, which restricts it to bounded offline use. An odometry also has to leave room for other software by keeping a small and predictable computational footprint. While related work shows promising accuracy, the hardware resources needed for these results are rarely reported.

\begin{figure}[!t]
    \centering
    \includegraphics[width=\linewidth]{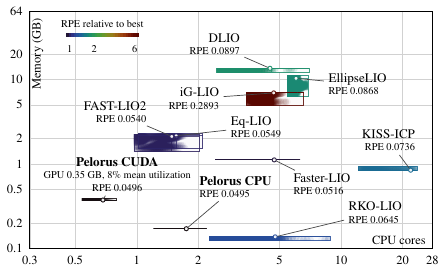}
    \caption{Memory and logical CPU cores of LIO systems on helipr/town02:ouster in logarithmic scale, measured on an Intel Core i7-14700 with an NVIDIA GeForce RTX 4060 Ti and bounded by its 28 logical cores and \SI{64}{\giga\byte}. Boxes span from the 50th percentile to the maximum with the measurement density inside, lines mark the 95th percentiles, and the box color encodes the relative pose error (RPE) as a multiple of the lowest one on a logarithmic scale. The RPE in the labels is the geometric mean of its translational and rotational parts. Pelorus achieves the lowest relative error with a small, bounded memory footprint. Only RKO-LIO uses less memory, but at a higher RPE and CPU load. With CUDA, Pelorus has the lowest CPU load.}

\label{fig:resource_usage}
\end{figure}

At the same time, embedded boards with a low power consumption increasingly feature general-purpose GPUs. Offloading LIO to these GPUs can significantly reduce its CPU footprint, but existing methods do not target these hardware limits. This work introduces Pelorus, a resource-aware LIO that achieves state-of-the-art relative accuracy within a bounded local map (\figref{resource_usage}) and scales down to an NVIDIA Jetson Orin NX, processing high-resolution scans at sensor rate.

Pelorus' main contributions are listed as follows:

\begin{enumerate}
  \item A fixed-capacity multiresolution local map of voxelized normal distributions with a relevance-based eviction score and a maturity-weighted update, which bounds memory independent of trajectory length while suppressing inconsistent observations.
  \item A \textit{resource-aware} LIO pipeline using preallocated buffers and data structures suited to single-instruction, multiple-thread (SIMT) execution to run at sensor rate on an embedded GPU.
  \item A permissive open-source implementation with CUDA, Vulkan, and Metal backends, distributed as libraries, framework integrations, and standalone binaries.\footnote{\coderepository}
\end{enumerate}

\section{Related Work}

Lidar-inertial odometry has to process millions of points per second online on the hardware of the platform it localizes. The memory of the local map must not grow with the trajectory for continuous operation, and the compute per scan must fit the sensor period on embedded processors while leaving room for the software that relies on the odometry. We review LIO methods along these two constraints.

Early methods like LOAM by Zhang and Singh~\cite{loam} and F-LOAM by Wang~et~al.~\cite{floam} reduce the compute by extracting edge and plane features before the registration, at the cost of assumptions on the scene. LOAM uses the IMU only for initial guesses and deskewing, which is known as \textit{loosely coupled} LIO, while F-LOAM works without an IMU. LIO-SAM by Shan~et~al.~\cite{liosam} couples IMU preintegration and lidar odometry factors in a factor graph that also estimates the IMU biases, a \textit{tightly coupled} formulation. The graph optimization is sequential which weighs on the ARM cores of embedded boards. Koide~et~al.~\cite{gicpgpu,glim} evaluate generalized iterative closest point (GICP) matching cost factors on the GPU, but the odometry of GLIM still bounds the optimization by a fixed-lag window of keyframes.

Filter-based methods avoid this cost by correcting the navigation state directly with the lidar measurements, so their resource use is dominated by the local map. FAST-LIO2 by Xu~et~al.~\cite{fastlio2} uses an iterated Kalman filter on manifolds~\cite{ikfom} and registers the scan points against an incremental kd-tree that is trimmed to a box around the sensor. Its memory is thus bounded by volume and depends on the point density, and rebuilding parts of the tree interrupts the otherwise per-point parallel neighbor search. Point-LIO by He~et~al.~\cite{pointlio} keeps this map and fuses every point individually into the filter state for robustness under highly dynamic motion. Faster-LIO by Bai~et~al.~\cite{fasterlio} replaces the tree with incremental voxels (iVox), a voxel hashmap that parallelizes neighbor search and map update and bounds the number of voxels with a least-recently-used (LRU) cache. Recency is a simple retention criterion, but it does not distinguish well-observed geometry from a few recent returns. DLIO by Chen~et~al.~\cite{dlio} instead registers the scan with GICP against a submap of keyframes after continuous-time motion correction. Since the keyframes are retained, its memory grows with the explored area. The Equivariant Filter (EqF) by van Goor~et~al.~\cite{eqf,eqftac} further evolves the state estimation model. It defines the estimation error through system symmetries, which reduces linearization errors. Tao~et~al.~\cite{eqlio} apply it to LIO in Eq-LIO, coupling navigation, IMU biases, and lidar--IMU extrinsics through a semidirect product symmetry with gravity constrained on the two-dimensional sphere.

Instead of storing points, a map can summarize each voxel by a distribution. This compresses the local geometry into a fixed size per voxel and lets every voxel be fitted and matched independently, which suits parallel hardware. The Normal Distributions Transform (NDT)~\cite{ndtbiber} fits a Gaussian per voxel. LiTAMIN by Yokozuka~et~al.~\cite{litamin} stabilizes NDT-based lidar odometry with a regularized covariance, and S-NDT by Renaut~et~al.~\cite{smoothndt} smooths the NDT map to widen the convergence basin. To reduce the computational load, Sun~et~al.~\cite{lightweightlio} combine NDT matching with IMU-based deskewing, and Ren~et~al.~\cite{odomfeat} extract plane and edge features before the NDT matching. iG-LIO by Chen~et~al.~\cite{chen2024iglio} tightly couples GICP with the IMU on an LRU-bounded voxel map like iVox and estimates a covariance for every scan point from neighboring voxels, so covariances are computed for both the map and the scan. FAR-LIO by Leitenstern~et~al.~\cite{farlio} accelerates a sparsity-aware GICP on a GPU voxel hashmap for autonomous racing and fuses it with the IMU in an extended Kalman filter with a kinematic motion model. These works reduce the cost per scan but mainly focus on runtime speed.

Uneven spatial sampling further motivates several map resolutions. MARS by Quenzel and Behnke~\cite{mars} combines multiresolution surfel maps with continuous-time registration and bounds each level by an egocentric volume and a keyframe window. VoxelMap by Yuan~et~al.~\cite{voxelmap} refines voxels in octrees according to the local planarity and keeps a global map. Pelorus adopts a multiresolution map of distributions, but bounds the number of distributions per level instead of the covered volume and selects the ones to keep by their relevance.

Hardware acceleration on embedded systems is a further strategy, for example truncated signed distance field (TSDF) mapping accelerated on a field-programmable gate array (FPGA)~\cite{hatsdf}, a heterogeneous FPGA and GPU SLAM for drones~\cite{redrose}, and FeatSense by Gaal~et~al.~\cite{featsense}, which runs feature-based odometry on the CPU and TSDF mapping on the GPU of Jetson boards. These systems lower the compute per scan, but their dense maps remain hard to bound in memory~\cite{featsense,redrose}.

In summary, existing LIO methods bound their local map by volume, by recency, or by a keyframe window, or not at all, and the methods that accelerate LIO on embedded hardware target the compute per scan. The resources the methods need on such hardware are rarely reported. To our knowledge, no LIO combines a tightly coupled filter with a map of distributions, retained by relevance, in a pipeline that processes high-resolution scans at sensor rate on an embedded GPU with a measured resource footprint. Pelorus addresses this combination.

\section{Method}
We call an odometry \textit{resource-aware} if its memory and compute footprints are bounded, small, and predictable, so it can run alongside the application that relies on it. Every step of Pelorus is therefore designed GPU-first with a focus on parallelization, cache locality, preallocation, and avoids data structures that grow with the input size.

\subsection{Preprocessing}
Vendor-specific input points are converted on the fly to a standardized point type by a per-point abstraction passed through the pipeline, which saves a temporary cloud buffer and a separate conversion pass. Every buffer, including scratch buffers for matrix operations, is sized by the scan and allocated once at initialization, so the pipeline does not allocate memory over time and keeps its accesses cache-local.

The first preprocessing step removes invalid measurements, such as returns from the platform itself, and downsamples the scan to harmonize the point density. Preprocessing runs sequentially or in parallel on the CPU, or on a GPU backend, where it requires an additional copy of the incoming cloud.

The IMU timestamps are then used to predict the state of an equivariant filter following Eq-LIO~\cite{eqlio}. The predictions are computed sequentially, stored, and used to deskew the scan in parallel.

\subsection{Equivariant State Estimation}

After predicting the state in the preprocessing step, the scan corrects the filter state. The 24-dimensional error state contains position, orientation, velocity, gyroscope and accelerometer biases, lidar--IMU rotation and translation, and a three-dimensional gravity vector. The extrinsic blocks stay inactive unless online extrinsic refinement is enabled. The correction uses a global-frame point-to-plane Jacobian and an iterated information-form update. Gravity is represented in $\mathbb{R}^3$ with a small initial covariance, whereas Eq-LIO constrains its direction on the two-dimensional sphere.

To keep the position error local over long trajectories, Pelorus uses a moving anchor $\mathbf{a}$. Let $\mathbf{R}$, $\mathbf{p}$, and $\mathbf{v}$ denote IMU orientation, position, and velocity in the world frame, and let $\mathbf{E}=\exp([\delta\boldsymbol{\theta}]_{\times})$ be the rotation induced by the orientation correction. The navigation state is updated as
$$
\begin{aligned}
\mathbf{R}^{+} &= \mathbf{E}\mathbf{R}, \\
\mathbf{p}^{+} &= \mathbf{a}+\delta\mathbf{p}+\mathbf{E}(\mathbf{p}-\mathbf{a}), \\
\mathbf{v}^{+} &= \delta\mathbf{v}+\mathbf{E}\mathbf{v}.
\end{aligned}
$$
Here $[\cdot]_{\times}$ denotes the skew-symmetric matrix. The anchor is set to the predicted position before scan correction, keeping the rotational moment arms local to the platform. When the anchor moves, the covariance is mapped to the new error coordinates with the Jacobian of this change. After each correction, it is reset to the corrected state, as in error-state filters.

\subsection{Multiresolution Local Map and Scan Registration}

The deskewed scan is registered against a local map of voxelized normal distributions, following the map representation proposed by Biber and Strasser~\cite{ndtbiber} and extended to 3-D by Magnusson~\cite{ndtbible}. Multiresolution maps provide geometric support at different spatial scales~\cite{mars,voxelmap}. Pelorus uses four hash levels with resolutions $\{0.5r,r,2r,4r\}$, where $r$ is the nominal voxel resolution. Each level receives one quarter of the configured total capacity $|M|$, up to rounding, and uses the same maturity-weighted update and relevance-based eviction described below.

The finest distributions are built from the current scan, and coarser distributions pool its child voxels so each observation contributes once per level. On the GPU backends, each level stores its distributions in a buffer sorted by voxel coordinate, and a hashmap with linear probing holds indices into this buffer for SIMT-friendly lookups. The CUDA, Vulkan, and Metal backends perform the same computations as the CPU implementation and differ from it only by floating-point rounding.

A query visits the levels from fine to coarse. On each visited level, the voxel containing the query point and its 26 surrounding voxels are searched, and the existing distributions in this vicinity are pooled into a single distribution, weighting their means and covariances by their sample counts with Chan's parallel update algorithm~\cite{chan1982updating}.

Planar feature extraction applies a principal component analysis (PCA) to the pooled covariance. For ordered eigenvalues $\lambda_0 \le \lambda_1 \le \lambda_2$, candidates require at least six supporting points, spread in both tangent directions. Association is restricted to the observed tangent extents of the patch. The query stops at the first level with a valid candidate, so each scan point contributes at most one point-to-plane residual to the filter correction. As in the implementation of FAST-LIO2~\cite{fastlio2}, a range-dependent gate rejects residuals that are large for the range of their point. Fine levels retain local detail, while coarse levels provide support where the finer levels are sparse or degenerate.

\subsection{Map Update}
After registration, the deskewed scan is transformed with the corrected pose and added to every level of the local map, voxel by voxel. If the voxel does not exist in the level, its distribution is inserted as a new entry. When the level is full, new entries compete with the existing ones in the eviction below, so its memory is bounded independent of trajectory length.

If the voxel already exists, Pelorus weights the incoming distribution $q$ by how well its mean fits the stored distribution $m$, measured by the regularized squared Mahalanobis distance
$$
D^2 = (\boldsymbol{\mu}_q - \boldsymbol{\mu}_m)^T (\mathbf{\Sigma}_m + \varepsilon\mathbf{I})^{-1} (\boldsymbol{\mu}_q - \boldsymbol{\mu}_m).
$$
Since $D^2$ is normalized by the voxel covariance, it is independent of the voxel size and the sensor. The weight
$$
\alpha = \exp\left(-\frac{\max(1, \ln(N))\, D^2}{2\kappa}\right)
$$
is the weight function of the Welsch M-estimator~\cite{welsch} with the scale $\kappa$, narrowed by the maturity $\max(1, \ln N)$ of the voxel, where $N$ is the number of points merged into it. Analogous to the pooling in the registration step, the voxel takes the mean and covariance of the mixture of both distributions with the weights $1-\alpha$ and $\alpha$,
$$
\begin{aligned}
\boldsymbol{\mu}_m' &= (1-\alpha)\,\boldsymbol{\mu}_m + \alpha\,\boldsymbol{\mu}_q, \\[1.0ex]
\mathbf{\Sigma}_m' &= (1-\alpha)\,\mathbf{\Sigma}_m + \alpha\,\mathbf{\Sigma}_q \\
&\quad + \alpha(1-\alpha)(\boldsymbol{\mu}_q - \boldsymbol{\mu}_m)(\boldsymbol{\mu}_q - \boldsymbol{\mu}_m)^T,
\end{aligned}
$$
and $N$ is increased by the number of incoming points. Pelorus thereby weights each observation by its consistency with the stored geometry instead of its point count. An observation at the stored mean ($D^2 = 0$) receives $\alpha = 1$ and replaces the distribution, so the map follows consistent observations. Observations that disagree with a mature voxel, caused by measurement outliers, dynamic objects, or residual pose errors, receive a small $\alpha$ and barely change it. Pelorus thus keeps established geometry stable and reduces the accumulated drift. Updates with $\alpha < 10^{-5}$ are skipped but still mark the voxel as observed. We choose $\kappa$ empirically in Section~\ref{sec:ablation_update}.

\subsection{Eviction}

After a scan has been merged into a level, the entries exceeding the level's capacity are evicted by the score ${S = T/P}$, where $T$ is the number of scans since the voxel was last updated and $P$ the number of points fused into the distribution. A high $S$ marks voxels that are either sparsely observed, such as transient returns from dynamic objects, or no longer observed, such as structures the sensor has left behind. In contrast to pure recency (LRU), a voxel with many accumulated points survives longer without updates, so stable structure remains in the map during short occlusions. Voxels observed in the current scan have $T=0$ and are only evicted after all stale ones. Exactly the excess entries with the highest $S$ are removed. Since each level is evicted against its own share of the capacity, coarse levels cannot displace fine detail and the total map size is bounded by $|M|$ after every update for any trajectory length.

\section{Experiments}

To evaluate the contributions of Pelorus, we measured odometry accuracy, latency, and resource footprints of the proposed methods on the Newer College~\cite{newercollege}, HeLiPR~\cite{helipr}, and Oxford Spires~\cite{oxfordspires} datasets. The complete overview is shown in Table~\ref{tab:accuracy_metrics}. In the following, the Newer College sequence park\_merged:os0\_128 is abbreviated as NC, the Oxford Spires sequences observatory\_quarter\_02 and christ\_church\_02 as OQ02 and CC02, and the HeLiPR sequences with the Ouster OS2 by their names bridge02 and town02. As is common for odometry, we use the relative pose error (RPE) at $\delta=\SI{1}{\second}$ as the primary accuracy metric~\cite{oxfordspires}. An odometry frontend has to deliver accurate motion increments, while global consistency is usually restored by a SLAM backend through loop closures and pose-graph optimization. In submap-based SLAM, the odometry determines the internal geometry of each submap, and the pose graph only corrects the relative placement of the submaps~\cite{cartographer}. The relative accuracy of the frontend therefore directly limits the quality of the map.

A low absolute pose error (APE) does not imply accurate motion increments, therefore the APE is reported for completeness only. On OQ02, iG-LIO achieves nearly half the APE of Pelorus at a four times higher translational RPE. The APE depends on when and in which direction the errors occur, since an early rotation error displaces the whole remaining trajectory while opposite errors cancel after the alignment~\cite{kuemmerle2009measuring,zhang2018tutorial}. Each trajectory is first normalized to start in the origin and then fitted to the reference with Umeyama alignment~\cite{umeyama}. All trajectories are measured using the respective reference implementations built with the Robot Operating System (ROS) framework. For the comparison, each method is run with real-time playback and reliable quality of service to guarantee each message being processed in the correct order and in full when the method is finished. For testing the methods on power-constrained devices, the NVIDIA Jetson board is also used as an evaluation platform. The bags are streamed over the network with \playbacktool{},\playbacktoolnote{} since \texttt{ros2 bag play} did not deliver all messages reliably in this setup. We compare against FAST-LIO2~\cite{fastlio2}, Eq-LIO~\cite{eqlio}, Faster-LIO~\cite{fasterlio}, RKO-LIO~\cite{rkolio}, EllipseLIO~\cite{border2026ellipselio}, DLIO~\cite{dlio}, iG-LIO~\cite{chen2024iglio}, and the lidar-only KISS-ICP~\cite{kissicp}. Pelorus achieves the lowest RPE in seven of the ten metrics. On bridge02, FAST-LIO2 and Faster-LIO are slightly more accurate in translation and rotation. On OQ02, Faster-LIO has a lower rotational RPE.

Pelorus does not aim for the lowest accumulated drift. On the long HeLiPR sequences, FAST-LIO2 and Faster-LIO reach a lower APE, which is dominated by slowly accumulating, mostly vertical drift over kilometers. Pelorus discards geometry the sensor has left behind to keep its memory bounded and small, so this geometry no longer supports the registration when the platform returns. The capacity sweep in \secref{map_capacity} shows that this is a deliberate trade-off. Beyond the chosen capacity, a larger map still lowers the APE but no longer the RPE. Pelorus therefore aims for the operating point at which the local accuracy saturates within the resources of the target hardware, and leaves global consistency to a backend. For this purpose, the implementation provides an odometry module for the GLIM framework~\cite{glim}, whose submap-based global optimization corrects the accumulated drift. None of the compared methods is globally consistent on its own. All methods that finish bridge02 accumulate between 14 and \SI{49}{\metre} of APE over its \SI{14.6}{\km}.  EllipseLIO retains every map point, allowing revisited geometry to constrain drift, but its growing map overflows its octree after \SI{3.7}{\km} of bridge02. Eq-LIO diverges on bridge02 after \SI{35}{\minute}.

Additionally, Surfel-LIO~\cite{surfellio} has been tested due to method similarity but it did not finish the HeLiPR datasets and achieved a high, out-of-distribution error for Newer College and is therefore not shown in the table. Similarly, FAR-LIO~\cite{farlio} has been evaluated as a recent GPU-accelerated method but did not successfully finish any of the tested datasets and is therefore also excluded from the table.

\begin{figure}[!t]
    \centering
    \includegraphics[width=\linewidth]{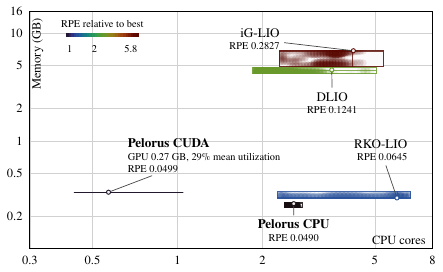}
    \caption{Resource footprints on an NVIDIA Jetson Orin NX running helipr/town02:ouster in real time, with the axes bounded by its eight CPU cores and \SI{16}{\giga\byte}. With CUDA, Pelorus handles the high-bandwidth OS2 lidar at sensor rate, and both builds hold the desktop accuracy with a fraction of the resources of the other methods. FAST-LIO2, Faster-LIO, and KISS-ICP drop too many scans and lose track, and EllipseLIO and Eq-LIO crash on the board, so they are not shown. Table~\ref{tab:nx_timings} lists all methods.}
    \label{fig:resource_usage_nx}
\end{figure}

The resource footprints are measured on a desktop with an Intel Core i7-14700 (28 logical cores, \SI{64}{\giga\byte}) and an NVIDIA GeForce RTX 4060 Ti, where all methods process the scans faster than sensor frequency (\figref{resource_usage}), and on an NVIDIA Jetson Orin NX with eight cores at \SI{1.98}{\giga\hertz}, a GPU at \SI{918}{\mega\hertz} and \SI{16}{\giga\byte} of shared memory, representing the compute of battery-powered systems (\figref{resource_usage_nx}). The HeLiPR sequences use an Ouster OS2-128 with up to \SI{400}{\metre} range and $1.3$ million points per second. This showcases a high-bandwidth sensor processed on a compute-constrained board.

Unless stated otherwise, Pelorus uses one configuration for all datasets: a nominal voxel resolution $r$ of \SI{0.5}{\metre}, the voxel size of the baselines, resulting in the four map resolutions $\{0.5r,r,2r,4r\}$, an input downsampled to half the nominal resolution, a total map capacity of $|M| = \num{131072}$ distributions, a sensitivity of $\kappa = 10$, at most four equivariant-filter correction iterations and a budget of four CPU threads. Only sensor- and platform-specific settings differ, such as the extrinsics and a minimum point range of \SI{4}{\metre} (\SI{1}{\metre} for Oxford Spires). To compare all methods under the same conditions, we ported Faster-LIO and Eq-LIO to ROS~2 and patched the other reference implementations. FAST-LIO2 receives the recorded point timestamps of the Hesai lidar in Oxford Spires instead of estimating them from the azimuth. The covariance regularization $\mathbf{U}\mathbf{D}\mathbf{V}^T$ of iG-LIO, which inverts the normal direction of a nearly planar neighborhood when its smallest eigenvalue is numerically negative and drove the filter state of iG-LIO to NaN within seconds on this dataset, is replaced by $\mathbf{U}\mathbf{D}\mathbf{U}^T$. Since the handheld Oxford Spires sequences start while the platform rotates, Faster-LIO and iG-LIO initialize their IMU biases with zero on them. KISS-ICP registers each scan without deskewing, so its pose refers to the scan center, and is stamped accordingly instead of with the scan start. Map saving of FAST-LIO2 is disabled so that its memory is not inflated. Eq-LIO is built on the code of FAST-LIO2 and uses its settings, so the two differ only in the filter. Apart from these changes, the sensor settings and a voxel size of \SI{0.5}{\metre}, the baselines keep the parameters of their authors. All configurations are published in the Pelorus repository.

Most compared methods do not bound their memory. DLIO, for example, accumulates keyframes, which eventually exceeds the memory of the device. Pelorus finishes the \SI{8.2}{\km} town02 sequence on both systems at sensor rate with a small, bounded memory and CPU footprint.

With GPU offloading, the CPU load decreases further, as shown in \figref{resource_usage_nx} for the Jetson Orin NX. Methods whose processing time exceeds the \SI{100}{\ms} lidar period lose scans and accuracy. FAST-LIO2, Faster-LIO, and KISS-ICP lose track on town02, although they finish it on the desktop, and Eq-LIO diverges after \SI{110}{\s} of the sequence and crashes.

Table~\ref{tab:nx_timings} lists the share of processed scans and the 95th-percentile (p95) latency of all methods and the p95 processing time of the Pelorus configurations, with $_{\times x}$ denoting the budget of logical cores. Only the GPU configurations of Pelorus stay below the sensor period. On the CPU alone, Pelorus skips scans while its RPE stays at the desktop level. Preprocessing on four CPU cores is faster than on the GPU, which requires an additional copy of the cloud. Pelorus buffers at most eight scans between its pipeline stages and drops incoming scans when a buffer is full, so latency and memory stay bounded under overload. Processing time sums the compute time of all pipeline stages for one scan. The end-to-end latency additionally includes the queueing and the wait for IMU measurements that cover the scan, and is measured for the baselines from the reception of the scan in its callback to the published odometry. DLIO has a lower latency than Pelorus, at twice its RPE, and is the only other method that processes nearly all scans within this level of latency.

For the energy efficiency, Table~\ref{tab:nx_timings} reports the mean power at the board input. All methods draw between 11.4 and \SI{13.1}{\watt}, except KISS-ICP with \SI{15.7}{\watt}, so at a similar power draw, Pelorus is the only method that keeps the sensor rate and the accuracy of the desktop. To relate energy and accuracy in one number, we report the energy--error product E$\cdot$RPE, analogous to the energy--delay product of processors~\cite{edp}. It is the board energy of the run divided by the number of processed scans, multiplied by the translational RPE. A method that drops scans pays twice, through a higher energy per processed scan and through a higher error. With about \SI{1.2}{\joule} per scan, Pelorus needs as much energy as the other methods that keep the sensor rate, and all its configurations reach a lower E$\cdot$RPE than every baseline. RKO-LIO has a \SI{22}{\percent} higher E$\cdot$RPE, DLIO twice, and iG-LIO more than five times the E$\cdot$RPE of Pelorus.

\begin{table*}[t]
\small
\setlength{\tabcolsep}{2.5pt}
\centering
\sisetup{
    table-format=2.4, 
    table-number-alignment=center,
    detect-weight=true,
    detect-shape=true,
    input-ignore={\best,\second}
}
\caption{Accuracy Across LIO Systems}
\label{tab:accuracy_metrics}
\medskip

\resizebox{\textwidth}{!}{%
\begin{tabular}{l l *{9}{S}}
\toprule
\textbf{Dataset}
  & \textbf{Metric}
& {\textbf{Pelorus}}
& {\shortstack{\textbf{FAST-LIO2}\\\cite{fastlio2}}}
& {\shortstack{\textbf{Eq-LIO}\\\cite{eqlio}}}
& {\shortstack{\textbf{Faster-LIO}\\\cite{fasterlio}}}
& {\shortstack{\textbf{RKO-LIO}\\\cite{rkolio}}}
& {\shortstack{\textbf{EllipseLIO}\\\cite{border2026ellipselio}}}
& {\shortstack{\textbf{DLIO}\\\cite{dlio}}}
& {\shortstack{\textbf{iG-LIO}\\\cite{chen2024iglio}}}
& {\shortstack{\textbf{KISS-ICP}\\\cite{kissicp}}} \\
\midrule
  Newer College \cite{newercollege} & RPE T
    & \best{0.0255} & 0.0373 & 0.0735 & 0.0310 & 0.0496 & 0.0498 & 0.0529 & \second{0.0276} & 0.2468 \\
  park\_merged:os0\_128 & RPE R
    & \best{0.2821} & 0.4795 & 0.5227 & 0.4815 & 0.4281 & 0.4762 & 1.3187 & \second{0.2913} & 0.8397 \\
  \SI[parse-numbers=false]{2.4}{\kilo\metre} & \italtxt{APE T}
    & \italtxt{0.2925} & \italtxt{0.2655} & \italtxt{11.4013} & \italsecond{0.2134} & \italtxt{1.4669} & \italtxt{0.2970} & \italtxt{0.2749} & \italbest{0.1973} & \italtxt{0.7702} \\
   & \italtxt{APE R}
    & \italtxt{0.6632} & \italtxt{0.6196} & \italtxt{5.8040} & \italsecond{0.5851} & \italtxt{1.0732} & \italtxt{0.6357} & \italtxt{1.3189} & \italbest{0.4782} & \italtxt{1.0919} \\
  \midrule
  HeLiPR \cite{helipr} & RPE T
    & 0.0648 & \best{0.0611} & \DNF & \second{0.0628} & 0.0907 & \DNF & 0.0865 & 0.2487 & 0.1302 \\
  bridge02:ouster & RPE R
    & 0.0568 & \second{0.0536} & \DNF & \best{0.0437} & 0.0582 & \DNF & 0.0826 & 0.1230 & 0.0689 \\
  \SI[parse-numbers=false]{14.6}{\kilo\metre} & \italtxt{APE T}
    & \italtxt{45.0027} & \italbest{14.2422} & \DNFit & \italtxt{15.9210} & \italtxt{31.7560} & \DNFit & \italsecond{14.9947} & \italtxt{26.4636} & \italtxt{49.1110} \\
   & \italtxt{APE R}
    & \italtxt{5.6010} & \italsecond{2.8695} & \DNFit & \italbest{2.5997} & \italtxt{3.7432} & \DNFit & \italtxt{3.3853} & \italtxt{3.6126} & \italtxt{3.8674} \\
  \midrule
  HeLiPR \cite{helipr} & RPE T
    & \best{0.0391} & 0.0424 & 0.0422 & \second{0.0404} & 0.0454 & 0.0845 & 0.0651 & 0.2266 & 0.0489 \\
  town02:ouster & RPE R
    & \best{0.0629} & 0.0688 & 0.0713 & \second{0.0658} & 0.0916 & 0.0892 & 0.1237 & 0.3693 & 0.1109 \\
  \SI[parse-numbers=false]{8.2}{\kilo\metre} & \italtxt{APE T}
    & \italtxt{5.5395} & \italsecond{2.6686} & \italtxt{5.8057} & \italbest{2.4071} & \italtxt{5.2028} & \italtxt{2.6985} & \italtxt{3.5233} & \italtxt{3.2817} & \italtxt{7.4419} \\
   & \italtxt{APE R}
    & \italtxt{1.6773} & \italtxt{1.1031} & \italtxt{1.4802} & \italbest{0.9691} & \italtxt{1.3872} & \italtxt{1.0415} & \italtxt{1.3900} & \italsecond{0.9946} & \italtxt{1.5207} \\
  \midrule
  Oxford Spires \cite{oxfordspires} & RPE T
    & \best{0.0073} & 0.0089 & 0.0104 & \second{0.0085} & 0.0491 & 0.0345 & 0.0282 & 0.0400 & 0.1542 \\
  observatory\_quarter\_02 & RPE R
    & \second{0.1357} & 0.1364 & 0.1398 & \best{0.1268} & 0.4929 & 0.2050 & 1.1193 & 0.1400 & 0.5868 \\
  \SI[parse-numbers=false]{0.4}{\kilo\metre} & \italtxt{APE T}
    & \italtxt{0.0971} & \italtxt{0.0639} & \italtxt{0.0945} & \italtxt{0.0584} & \italtxt{0.0567} & \italsecond{0.0480} & \italtxt{0.0758} & \italbest{0.0424} & \italtxt{0.4532} \\
   & \italtxt{APE R}
    & \italtxt{0.3298} & \italtxt{0.2677} & \italtxt{0.3566} & \italsecond{0.2368} & \italtxt{0.4112} & \italtxt{0.2374} & \italtxt{0.9165} & \italbest{0.1951} & \italtxt{0.7743} \\
  \midrule
  Oxford Spires \cite{oxfordspires} & RPE T
    & \best{0.0071} & 0.0096 & \second{0.0092} & 0.0102 & 0.0594 & 0.0484 & 0.0314 & 0.0308 & 0.1489 \\
  christ\_church\_02 & RPE R
    & \best{0.1237} & 0.1339 & \second{0.1332} & 0.1400 & 0.4573 & 0.2037 & 1.0385 & 0.2041 & 1.0148 \\
  \SI[parse-numbers=false]{0.6}{\kilo\metre} & \italtxt{APE T}
    & \italsecond{0.0889} & \italtxt{0.3592} & \italtxt{0.3494} & \italtxt{0.3245} & \italtxt{0.4127} & \italtxt{0.1846} & \italbest{0.0861} & \italtxt{0.5423} & \italtxt{0.4933} \\
   & \italtxt{APE R}
    & \italbest{0.3240} & \italtxt{0.5690} & \italtxt{0.5728} & \italsecond{0.5115} & \italtxt{1.3449} & \italtxt{0.8581} & \italtxt{0.9626} & \italtxt{1.0125} & \italtxt{1.6444} \\
\bottomrule
\end{tabular}
}

\smallskip
\begin{minipage}{\linewidth}
\footnotesize RPE ($\delta=\SI{1}{\second}$) and APE (root-mean-square error) are divided into
  translational (meters) and rotational (degrees) parts. Pelorus is evaluated without kinematic constraints and the CPU build. RPE is the primary metric for odometry~\cite{oxfordspires}. APE is shown for reference in \italtxt{gray} and not ranked. \textbf{Best} results are marked in bold, \underline{second best} are underlined. \textit{d.n.f.} marks a run that did not finish the sequence.
\end{minipage}

\end{table*}

\begin{table}[t]
\centering
\sisetup{
    table-format=2.4, 
    table-number-alignment=center,
    detect-weight=true,
    detect-shape=true,
    input-ignore={\best,\second}
}
\caption{Timings, Power, and Energy Efficiency on the Jetson Orin NX}
\footnotesize
\setlength{\tabcolsep}{3pt}
\begin{tabularx}{\linewidth}{Xrrrrr}
\toprule
\textbf{Method} & \textbf{Scans} & \textbf{Latency} & \textbf{Proc.} & \textbf{Power} & \textbf{E$\cdot$RPE} \\
 & (\%) & p95 (ms) & p95 (ms) & (W) & (mJ\,m) \\
\midrule
Pelorus Vulkan & 100.0 & 192 & 97.8 & 13.1 & 50.3 \\
Pelorus CUDA & 99.7 & 175 & 83.7 & 12.4 & 48.4 \\
Pelorus CUDA + CPU$_{\times4}$ & 100.0 & 173 & 78.1 & 12.2 & \textbf{46.8} \\
Pelorus CPU$_{\times4}$ & 87.6 & 3105 & 220.3 & 12.0 & 50.7 \\
Pelorus CPU$_{\times8}$ & 93.1 & 2616 & 173.2 & 12.4 & 49.9 \\
\midrule
FAST-LIO2 & 10.3 & $>10^6$ & -- & 11.8 & \num{54113} \\
Eq-LIO\textsuperscript{a} & \textit{d.n.f.} & \num{119775} & -- & 11.4 & -- \\
Faster-LIO & 90.1 & \num{83315} & -- & 12.2 & 5402 \\
RKO-LIO & 98.9 & 3176 & -- & 12.7 & 57.0 \\
EllipseLIO\textsuperscript{a} & \textit{d.n.f.} & 178 & -- & 11.9 & -- \\
DLIO & 99.9 & 119 & -- & 12.4 & 96.2 \\
iG-LIO & 100.0 & 1815 & -- & 11.8 & 260.2 \\
KISS-ICP & 40.8 & 554 & -- & 15.7 & \num{11317} \\
\bottomrule
\end{tabularx}

\smallskip
\begin{minipage}{\linewidth}
\footnotesize Measured on helipr/town02:ouster. Mean power at the board input. The idle board draws \SI{9.3}{\watt}. E$\cdot$RPE is the board energy per processed scan times the translational RPE, lower is better.\\\textsuperscript{a}Until the crash.
\end{minipage}
\label{tab:nx_timings}
\end{table}

\subsection{Map Capacity}
\label{sec:map_capacity}
\begin{figure}[!t]
    \centering
    \includegraphics[width=\linewidth]{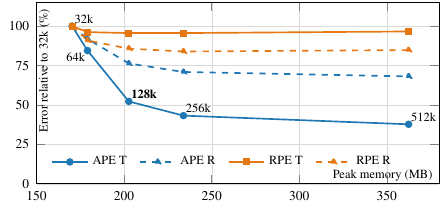}
    \caption{Capacity sweep on helipr/town02:ouster with the CPU build. Translational and rotational APE and RPE over the peak resident memory for map capacities $|M|$ from 32k to 512k distributions (k~=~1024), each relative to the smallest capacity. The capacity used is marked in bold.}
    \label{fig:capacity_sweep}
\end{figure}

The map capacity $|M|$ is the main resource parameter of Pelorus. \figref{capacity_sweep} shows the accuracy over the peak memory on town02 for capacities from \num{32768} to \num{524288} distributions, each measured in a single run of the CPU build. The relative error saturates at $|M| = \num{131072}$. Larger maps change the RPE by less than \SI{0.001}{\metre} and \SI{0.002}{\degree}, while they add \SI{31}{\mega\byte} at twice and \SI{160}{\mega\byte} at four times the capacity. The absolute error keeps decreasing with the capacity, from 10.6 to \SI{4.0}{\metre}, as larger maps retain older geometry that constrains the accumulated drift. The CPU load stays between 2.2 and 2.7 logical cores (p95) for all capacities. Beyond $|M| = \num{131072}$, the value used for all datasets, additional memory therefore only lowers the accumulated drift, which a backend corrects anyway, so this capacity marks the operating point on this hardware budget.

\subsection{Ablation: Eviction Strategy}
To isolate the relevance-based eviction, we replace the score ${S = T/P}$ with a least-recently-updated (LRU) order, i.e., ${S = T}$, behind a compile-time flag and compare two otherwise identical CPU builds. We reduce the capacity to three values at which eviction is active throughout the sequences. Table~\ref{tab:ablation_eviction} shows that the relative pose error is nearly unchanged. On the three handheld sequences, where the platform often returns to recently mapped structure, keeping well-observed distributions lowers the absolute translation error by \SIrange{21}{72}{\percent} in eight of nine cases, because this structure is still in the map when the platform returns. The exception is OQ02 at the smallest capacity, where both variants drift strongly. On town02, a drive with long stretches through unseen areas, recency alone is the better relevance measure and LRU achieves a lower absolute error at all three capacities. In real-time runs, its CPU load and peak memory exceeded those of LRU by at most \SI{5}{\percent}. The relevance score therefore helps when mapped areas are revisited within the lifetime of the local map, at no additional resource cost.

\begin{table}[t]
\centering
\caption{Eviction Ablation with Reduced Map Capacity}
\label{tab:ablation_eviction}
\footnotesize
\setlength{\tabcolsep}{2pt}
\begin{tabular}{@{}l l c c c c c@{}}
\toprule
 & & \multicolumn{4}{c}{\textbf{APE T} (m)} & \textbf{RPE T} \\
\cmidrule(lr){3-6}
$|M|$ & \textbf{Eviction} & \textbf{NC} & \textbf{OQ02} & \textbf{CC02} & \textbf{town02} & (\si{\percent}) \\
\midrule
\num{16384} & LRU & 5.36 & \best{0.728} & 0.411 & \best{16.0} & \\
 & $T/P$ & \best{2.15} & 0.759 & \best{0.226} & 18.2 & +2 \\
\midrule
\num{32768} & LRU & 1.88 & 0.404 & 0.187 & \best{9.16} & \\
 & $T/P$ & \best{1.32} & \best{0.181} & \best{0.132} & 9.75 & +1 \\
\midrule
\num{65536} & LRU & 2.10 & 0.365 & 0.217 & \best{6.01} & \\
 & $T/P$ & \best{0.591} & \best{0.289} & \best{0.133} & 7.49 & +1 \\
\bottomrule
\end{tabular}

\smallskip
\begin{minipage}{\linewidth}
\footnotesize Bold marks the lower APE T of each pair. RPE T is the mean change of $T/P$ relative to LRU.
\end{minipage}
\end{table}

\subsection{Ablation: Map Update}
\label{sec:ablation_update}
To isolate the maturity-weighted update, we replace it with the count-weighted pooling used for the registration neighborhood and vary the sensitivity $\kappa$ on the three handheld sequences, where the platform revisits mapped areas. The count-weighted update lets every observation shift the distributions, including those taken under residual pose errors. Table~\ref{tab:ablation_update} shows that $\kappa = 10$ gives the lowest absolute translation error on both Oxford Spires sequences and the largest mean reduction over the three sequences. A narrower kernel also rejects consistent observations, and wider kernels admit inconsistent ones. On Newer College, all variants lie within \SI{13}{\percent} of each other, so at this map capacity the update has little effect there. We use $\kappa = 10$ for all datasets. On Oxford Spires, the maturity-weighted update lowers the absolute translation error by \SI{57}{\percent} and \SI{29}{\percent} at an \SIrange{8}{11}{\percent} (\SIrange{0.5}{0.7}{\mm}) higher translational RPE, while on town02, a drive with few revisits, both update rules are within \SI{4}{\percent} of each other. The update therefore trades a small part of the relative accuracy for a lower drift where the platform revisits mapped areas. We accept this trade-off because the local map is reused whenever the platform returns, so distributions that follow every observation would pass residual pose errors on to later registrations. On Oxford Spires, the stable map already pays off over spans typical for submaps, with a \SI{19}{\mm} (\SI{14}{\percent} and \SI{17}{\percent}) lower translational RPE at $\delta = \SI{30}{\s}$.

\begin{table}[t]
\centering
\caption{Map Update Ablation}
\label{tab:ablation_update}
\footnotesize
\setlength{\tabcolsep}{2pt}
\begin{tabular}{@{}l c c c c c@{}}
\toprule
 & \multicolumn{3}{c}{\textbf{APE T} (m) and change (\si{\percent})} & \multicolumn{2}{c}{\textbf{Mean} (\si{\percent})} \\
\cmidrule(lr){2-4}\cmidrule(l){5-6}
\textbf{Update} & \textbf{NC} & \textbf{OQ02} & \textbf{CC02} & \textbf{APE T} & \textbf{RPE T} \\
\midrule
Count-weighted & 0.309 & 0.224 & 0.125 & 0 & 0 \\
$\kappa = 5$ & 0.308 (+0) & 0.123 ($-$45) & \second{0.101 ($-$19)} & \second{$-$22} & +5 \\
$\kappa = 10$ & \second{0.292 ($-$5)} & \best{0.097 ($-$57)} & \best{0.089 ($-$29)} & \best{$-$30} & +5 \\
$\kappa = 20$ & 0.317 (+3) & 0.120 ($-$46) & 0.102 ($-$19) & $-$21 & +7 \\
$\kappa = 40$ & 0.298 ($-$3) & 0.145 ($-$35) & 0.109 ($-$13) & $-$17 & +10 \\
$\kappa = 80$ & \best{0.280 ($-$9)} & \second{0.108 ($-$52)} & 0.140 (+12) & $-$17 & +11 \\
\bottomrule
\end{tabular}

\end{table}

\subsection{Ablation: Multiresolution Map}
The multiresolution map is adopted from prior work~\cite{mars,voxelmap}, so we verify that the accuracy of Pelorus does not rest on it. We replace the four levels by a single level with the same total capacity and the resolutions $r \in \{0.3, 0.5, 0.7\}\,\si{\metre}$, keeping all other components. Table~\ref{tab:ablation_multires} shows that the multiresolution map has the lowest translational RPE on all four sequences. A single level at the nominal resolution of \SI{0.5}{\metre}, the voxel size of the baselines, is between \SI{11}{\percent} and \SI{114}{\percent} less accurate. The resolution is not equivalent, since one distribution summarizes a whole voxel, whereas the baselines keep the downsampled points and fit planes to their neighbors. The best single resolution, \SI{0.3}{\metre}, comes within \SI{6}{\percent} on three sequences and would still rank first or tied in Table~\ref{tab:accuracy_metrics} on all four. The relative accuracy of Pelorus therefore does not rest on the multiresolution map itself. On CC02, however, the \SI{0.3}{\metre} level is \SI{22}{\percent} less accurate, so no single resolution matches the pyramid on all sequences. The pyramid reaches its accuracy without choosing a resolution per dataset, and in real-time runs on NC and town02 at a \SIrange{30}{35}{\percent} lower CPU load than the \SI{0.3}{\metre} level. Memory stays bounded by the same capacity in all configurations. Taken together, the eviction and the update bound the memory and reduce the accumulated drift at a small cost in relative accuracy. How the relative accuracy divides between map, filter, and deskewing remains open, but the filter alone does not explain it, since Eq-LIO is close to FAST-LIO2 on town02 and CC02 and less accurate elsewhere (Table~\ref{tab:accuracy_metrics}).

\begin{table}[t]
\centering
\caption{Multiresolution Ablation with Equal Total Capacity}
\label{tab:ablation_multires}
\footnotesize
\setlength{\tabcolsep}{2pt}
\begin{tabular}{@{}l c c c c@{}}
\toprule
 & \multicolumn{4}{c}{\textbf{RPE T} (m)} \\
\cmidrule(l){2-5}
\textbf{Map} & \textbf{NC} & \textbf{OQ02} & \textbf{CC02} & \textbf{town02} \\
\midrule
Multiresolution & \best{0.0255} & \best{0.0073} & \best{0.0071} & \best{0.0391} \\
Single \SI{0.3}{\metre} & \second{0.0261} & \second{0.0077} & \second{0.0087} & \second{0.0404} \\
Single \SI{0.5}{\metre} & 0.0284 & 0.0127 & 0.0153 & 0.0438 \\
Single \SI{0.7}{\metre} & 0.0330 & 0.0211 & 0.0383 & 0.0492 \\
\bottomrule
\end{tabular}

\end{table}

\section{Conclusion and Future Work}
\label{sec:conclusion}

We presented Pelorus, a resource-aware lidar-inertial odometry for the computational constraints of mobile platforms. A fixed-capacity multiresolution local map combines relevance-based eviction with maturity-weighted distribution updates to retain useful geometry within bounded memory. The parallel pipeline runs on multicore CPUs and embedded GPUs and achieves the best relative pose accuracy on most of the evaluated sequences with a small resource footprint. The open-source implementation provides CUDA, Vulkan, and Metal backends for deployment on these platforms.

Pelorus targets accurate incremental motion estimation under strict resource limits and leaves global consistency to a SLAM backend, as is common for odometry. Degeneracy-aware updates could make the accuracy on degenerate stretches at high speed (where all evaluated methods lose accuracy on bridge02) independent of the map resolution. Future work will further focus on extending this resource-aware architecture to full-scale SLAM. Specifically, we aim to investigate memory-bounded global pose graph optimization and compute-efficient loop closure detection. By adapting resource-awareness and Pelorus to the mapping task, we intend to provide a complete, long-term SLAM framework with small, predictable resource footprints for edge devices.

\bibliographystyle{IEEEtran}
{\raggedright
\bibliography{bibliography}
}

\flushcolsend
\end{document}